\documentclass[letterpaper, 10 pt, conference]{ieeeconf}  % Comment this line out if you need a4paper

\IEEEoverridecommandlockouts                              % This command is only needed if 
\usepackage{mathptmx} % assumes new font selection scheme installed
\usepackage{times} % assumes new font selection scheme installed
\usepackage{amsmath} % assumes amsmath package installed
\usepackage{multirow}
\usepackage[table]{xcolor}
\usepackage{tabularx}
\usepackage{epsfig, epsf}
\usepackage{url}
\usepackage{hyperref}

\title{\LARGE \bf
Multi-modal Panoramic 3D Outdoor Datasets for Place Categorization
\thanks{ {\bf This is the authors' manuscript. The final published article is
available at https://doi.org/10.1109/IROS.2016.7759669}}
\thanks{ {\bf Copyright 2016 IEEE. Personal use of this material is permitted. Permission
from IEEE must be obtained for all other uses, in any current or future
media, including reprinting/republishing this material for advertising
or promotional purposes, creating new collective works, for resale or
redistribution to servers or lists, or reuse of any copyrighted
component of this work in other works.}}
}

\author{Hojung Jung$^{1}$, Yuki Oto$^{1}$, Oscar M. Mozos$^{2}$, Yumi Iwashita$^{3}$ and Ryo Kurazume$^{3}$% <-this % stops a space
\thanks{$^{1}$Hojung Jung and Yuki Oto are with Graduate School of Information Science and Electrical Engineering, Kyushu University, Fukuoka 819-0395, Japan
        {\tt\small \{hojung,y\_oto\}@irvs.ait.kyushu-u.ac.jp}}%
\thanks{$^{2}$Oscar Martinez Mozos is with Polytechnic University of Cartagena (UPCT), Spain
        {\tt\small omozos@gmail.com}}
\thanks{$^{3}$Yumi Iwashita and Ryo Kurazume are with Faculty of Information Science and Electrical Engineering, Kyushu University, Fukuoka 819-0395, Japan
        {\tt\small \{yumi,kurazume\}@ait.kyushu-u.ac.jp}}
}

\begin{document}

\maketitle

\thispagestyle{empty}
\pagestyle{empty}

%%%%%%%%%%%%%%%%%%%%%%%%%%%%%%%%%%%%%%%%%%%%%%%%%%%%%%%%%%%%%%%%%%%%%%%%%%%%%%%%
\begin{abstract}

We propose two {Multi-modal Panoramic 3D Outdoor (MPO) Datasets} for place categorization with six semantic categories: forest, coast, residential area, urban area and indoor/outdoor parking lot. The first dataset consists of a high-resolution (9,000,000 points) color 3D point clouds with reflectance using a FARO laser scanner and synchronized color images, and 650 scans in total. The second dataset consists of a low-resolution (70,000 point) 3D point clouds with reflectance using a Velodyne laser scanner, and 34,200 scans in total. We conducted place categorization experiments of our datasets by evaluating standard feature descriptors and the highest classification results are 96.42\% and 89.67\% for each datasets. For benchmarking of outdoor place categorization, our datasets are publicly available on our website: \href{http://robotics.ait.kyushu-u.ac.jp/~kurazume/research-e.php?content=db-hidden}{http://robotics.ait.kyushu-u.ac.jp/}

\end{abstract}

%%%%%%%%%%%%%%%%%%%%%%%%%%%%%%%%%%%%%%%%%%%%%%%%%%%%%%%%%%%%%%%%%%%%%%%%%%%%%%%%

\section{INTRODUCTION}

%% Importance of Place categorization
Understanding the surrounding environment is an important research topic in robotics. Especially, one of the significant capabilities of mobile robots is knowing ``Where am I." This question means to identify a semantic category of location, called a place. The place can provide certain types of locations, such as `forest', `urban area', and `parking lot'. Information on the types of places can greatly improve the communication between robots and humans~\cite{zender2008ras, pronobis2012icra} and can allow robots to make decisions with context-based understanding when completing high-level tasks~\cite{stachniss2008amai}. Moreover, if a robot has the ability to categorize places, the robot will be able to properly execute a task even in unfamiliar surroundings.

\begin{figure}[thpb]
\centering
    \psfig{file=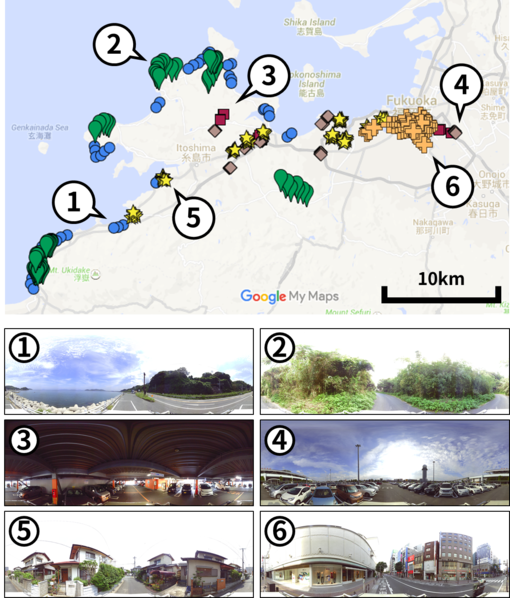,width=\columnwidth} 
\caption{\label{fig:map_velo} \small
  An example map of \href{http://robotics.ait.kyushu-u.ac.jp/~kurazume/research-e.php?content=db-hidden}{\it MPO Dataset} with six place categories: (1) forest, (2) coast, (3) indoor parking lot, (4) outdoor parking lot, (5) residential area and (6) urban area  
  }
\end{figure}

%% 2D place categorization
2D place categorization using 2D images has been achieved in high-level image understanding. 2D images can be captured by a camera or be collected by search engines, therefore, 2D image understanding and scene recognition are actively researched and evaluated by using several datasets, such as ImageNET~\cite{deng2009imagenet}, SUN database~\cite{xiao2010sun} and so on. ImageNET~\cite{deng2009imagenet} consists of a large collection of images with variability in visual appearance, which is available on the Internet using online search engines for each object category term. SUN database~\cite{xiao2010sun} used similar procedures to create place databases. The scene database contains 899 categories with 130,519 images of scenes and numerous state-of-the-art algorithms are validated for place categorization. However, those datasets do not provide 3D information.

%% 3D indoor place categorization
To extend the scope of 2D images, the 3D place categorization has been studied, which is mainly focused on indoor environment. Many indoor studies have been conducted using RGB-D datasets because the RGB-D sensor is a low-cost and easy to capture time series of depth image. It can capture each scene, which is a single video sequence consisting of multiple RGB-D frames. For those reason, the RGB-D indoor place datasets have become popular~\cite{Silberman:2012kh,Song:2015js}. The NYU-Depth V2 dataset~\cite{Silberman:2012kh} contains segmented 3D images of indoor environments, and consists of over 408,000 RGB-D images from 464 indoor scenes with per-frame accelerometer data. The 3DSUN RGB-D dataset~\cite{Song:2015js} is a 3D annotated benchmark scene containing 10,000 RGB-D images for 3D scene/place categorization and 3D reconstruction.

%% 3D outdoor dataset
Most studies on 3D place categorization are evaluated by indoor environment. Since the RGB-D sensor has a short detectable range, there is a lack of 3D place datasets especially for outdoor environment. As an 3D outdoor environment dataset, KITTI~\cite{geiger2012we} collected sequential 3D scan data using Velodyne HDL-64E laser scanner for visual SLAM with a low resolution. It contains four categories, which are city, residential, road and campus, and a purpose of the dataset is different from us, e.g., stereo, optical flow, visual odometry, 3D object detection and 3D tracking.

%% 3D outdoor place categorization
In this research, we present \href{http://robotics.ait.kyushu-u.ac.jp/~kurazume/research-e.php?content=db-hidden}{\it Dense and Sparse Multi-modal Panoramic 3D Outdoor (MPO) Datasets} for outdoor place categorization in static and dynamic environments for a mobile robot and autonomous vehicles. Our datasets are recorded by multi-modalities sensors, and provided as multi-resolution point clouds. Each dataset consists of six place categories including `forest', `coast', `indoor parking lot', `outdoor parking lot', `residential area' and `urban area'. The \href{http://robotics.ait.kyushu-u.ac.jp/~kurazume/research-e.php?content=db-hidden}{\it Dense MPO dataset} is recorded 650 color 3D scans using a high-resolution laser scanner. The \href{http://robotics.ait.kyushu-u.ac.jp/~kurazume/research-e.php?content=db-hidden}{\it Sparse MPO dataset} is recorded 5 hours long, 60 place scenarios and 34,200 scans with 40-60 kph velocity at 2-6 Hz using a real-time laser scanner, a 360 panoramic color camera, and a GPS. The proposed datasets are useful for benchmarking outdoor place categorization and publicly available on our website~\cite{MPO2016}. %The datasets can save your effort to create own dataset and encourage robotics researchers in fields of 3D visual learning and robot vision. 

%%%%%%%%%%%%%%%%%%%%%%%%%%%%%%%%%%%%%%%%%%%%%%%%%%%%%%%%%%%%%%%%%%%%%%%%%%%%%%%%

\section{DENSE MPO DATASETS}
The motivation of two \href{http://robotics.ait.kyushu-u.ac.jp/~kurazume/research-e.php?content=db-hidden}{\it MPO Datasets}~\cite{MPO2016} is for outdoor place categorization using 3D point clouds data. To create the datasets we utilize time-of-flight laser scanners of which an advantage is robust under various conditions, such as bright sunlight, darkness, and sudden illumination changes. In addition, the same semantic dataset with two different sensor configurations can be useful to evaluate a performance of outdoor place categorization. 

The \href{http://robotics.ait.kyushu-u.ac.jp/~kurazume/research-e.php?content=db-hidden}{\it Dense MPO Dataset} is composed of 3D high-resolution (5140 $\times$ 1757 pixels) color panoramic scans for outdoor place categorization. The main purpose is to create spatially synchronized color, reflectance, and range scan in static measurement condition. The dataset is able to utilize a set of three image modalities with six place categories and 650 scan sets in total. In this section, we will explain about sensors, experimental setup, place categories and data format of \href{http://robotics.ait.kyushu-u.ac.jp/~kurazume/research-e.php?content=db-hidden}{\it Dense MPO Dataset}.

\subsection{Data acquisition}    %DATA ACQUISITION
\begin{figure}[thpb]
\centering
    \psfig{file=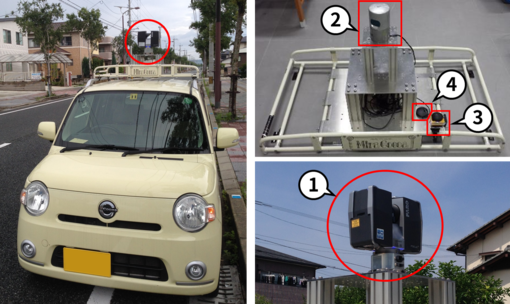, width=\columnwidth}
\caption{\label{fig:experiment} \small
  Experimental setup for \href{http://robotics.ait.kyushu-u.ac.jp/~kurazume/research-e.php?content=db-hidden}{\it Dense MPO Dataset} equipped with (1) a FARO Focus3D sensor system and for \href{http://robotics.ait.kyushu-u.ac.jp/~kurazume/research-e.php?content=db-hidden}{\it Sparse MPO Dataset} equipped with (2) a Velodyne HDL-32E laser scanner, (3) a Kodak PIXPRO SP360 camera and (4) a GARMIN GPS 18x LVC}
\end{figure}

In this dataset, we used a FARO Focus3D sensor system. The sensor includes a high-resolution time-of-flight laser scanner and a camera, so that it can provide range, reflectance and color panoramic images. The sensor system is installed on top of vehicle with 1.8 meters height, as shown in Fig.~\ref{fig:experiment}. The laser scanner rotates around the vertical axis to obtain a complete 360$^{\circ}$ panoramic color 3D point clouds around the vehicle. The system takes 3 minute to scan a set of synchronized laser scan and color image in total.

\subsubsection{FARO Focus3D laser scanner}
We use a laser scanner, which provides range data by measuring the round-trip time of a laser pulse reflected by an object. In addition to range data, the laser scanner can measure the strength of the reflected laser pulse, i.e., the reflectivity. The sensor can perform with a maximum range of 150 meters and a maximum angular resolution of 0.0009$^{\circ}$. Its field of view is $360^{\circ} \times 300^{\circ}$ (horizontal $\times$ vertical). In this experiment, we created 9 millions of points of a single scan with 0.07$^{\circ}$ and 0.17$^{\circ}$ of horizontal and vertical angular resolution for each. It took 1 minute and 30 seconds for a single scanning process.
%[CHECK] number of points!
%[CHECK] scanning time!

\subsubsection{FARO Focus3D camera}
In FARO Focus3D configuration, the system includes a camera which can capture the same direction of the laser scanner. After the FARO Focus3D rotates the laser beam around the vertical axis, the camera captures the same field of view images to obtain a synchronized color panoramic images with the range and reflectance scan, describing in Fig.~\ref{fig:experiment}. Each pair of a complete panoramic scan has a high resolution of 5140 $\times$ 1757 points. 

%\subsubsection{Experimental setup}
%We utilize a laser scanner and a camera. For our data acquisition, we installed a FARO Focus3D system on top of a mobile platform with 1.8 meters height, as shown in Fig.~\ref{fig:experiment}. A laser scanner rotates around the vertical axis to obtain a complete 360$^{\circ}$ panoramic color 3D point clouds around the mobile robot. The system takes 3 minute to scan a set of synchronized laser scan and color image in total.

\begin{table}
\centering
\footnotesize
%    \scriptsize
% \begin{center} 
\caption{
\small
DENSE MPO DATASET of outdoor scene containing 650 pairs of range, reflectance and color images
}
\label{tab:faro}
\begin{tabular}{|c|c|c|c|c|c|c|c|c|}    
    \hline
    \multirow{2}{*}{Category}    & \multicolumn{7}{|c|}{Number of scans by location} & \multirow{2}{*}{Total} \\ \cline{2-8}
       & {\tiny Set1}    & {\tiny Set2}    & {\tiny Set3}    & {\tiny Set4}    & {\tiny Set5}    & {\tiny Set6}    & {\tiny Set7}    & \\ \hline 
    \scriptsize
    Coast      & 14     & 14     & 16     & 12     & 17      & 14    & 16    & 103    \\ \hline
    Forest      & 16     & 16     & 17     & 18     & 16      & 16    & 17    & 116    \\ \hline
    Parking lot (in)      & 16     & 16     & 13     & 15     & 17      & 13    & 13    & 105    \\ \hline
    Parking lot (out)      & 15     & 17     & 16     & 15     & 15      & 14    & 16    & 108    \\ \hline
    Residential area      & 14     & 16     & 14     & 15     & 16      & 15    & 16    & 106    \\ \hline
    Urban area      & 16     & 17     & 16     & 16     & 15      & 16    & 16    & 112    \\ \hline
    \multicolumn{8}{|c|}{Total number of images}                 & 650   \\ \hline
\end{tabular}
% \end{center}
\vspace{-3mm}
\end{table}

\subsection{Dataset descriptions}    % DATA DESCRIPTIONS

\subsubsection{Place category}

\begin{figure}[thpb]
\centering
    \begin{tabular}{|c|c|}     \hline
        \multicolumn{2}{|c|}{\cellcolor[HTML]{EFEFEF}Range, reflectance and color panoramic images}    \\ \hline
        
        Forest    &     Coast \\ \hline
        \psfig{file=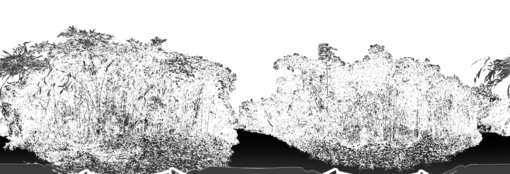,width= 0.45\columnwidth}
        & \psfig{file=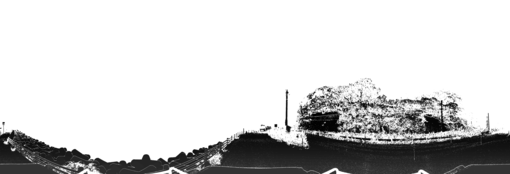,width= 0.45\columnwidth}  \\
        \psfig{file=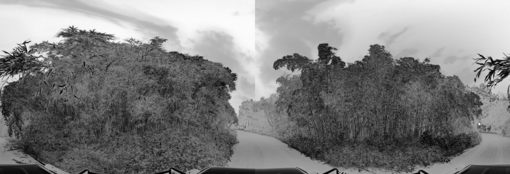,width= 0.45\columnwidth}
        & \psfig{file=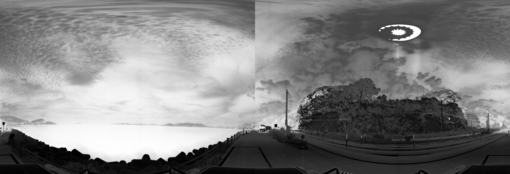,width= 0.45\columnwidth}  \\
        \psfig{file=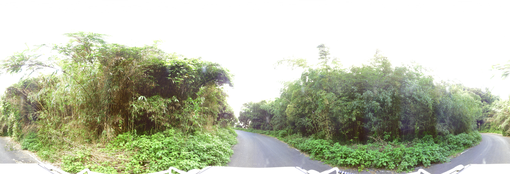,width= 0.45\columnwidth}
        & \psfig{file=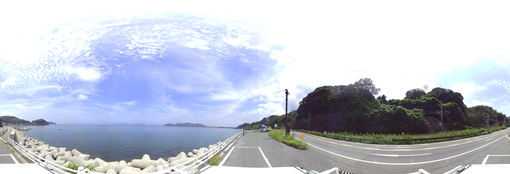,width= 0.45\columnwidth}    \\ \hline
        Parking lot (indoor)    & Parking lot (outdoor) \\ \hline
        \psfig{file=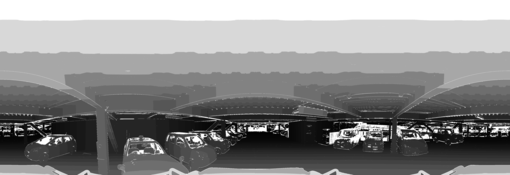,width= 0.45\columnwidth}
        & \psfig{file=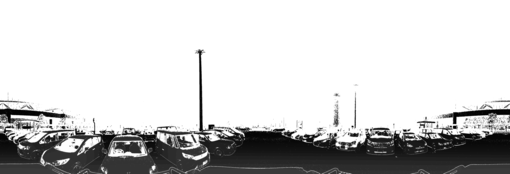,width= 0.45\columnwidth}  \\
        \psfig{file=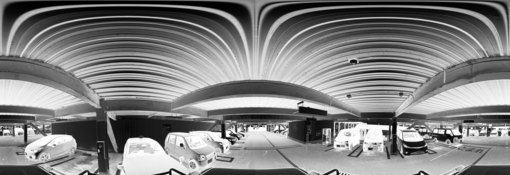,width= 0.45\columnwidth}
        & \psfig{file=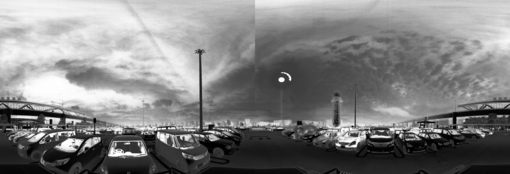,width= 0.45\columnwidth}  \\
        \psfig{file=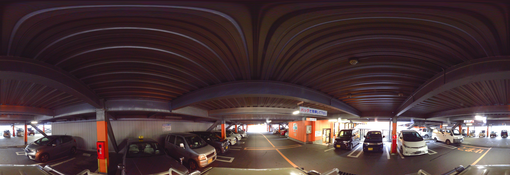,width= 0.45\columnwidth}         
        & \psfig{file=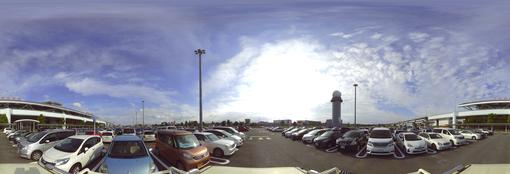,width= 0.45\columnwidth}    \\ \hline
        Residential area    & Urban area  \\ \hline
        \psfig{file=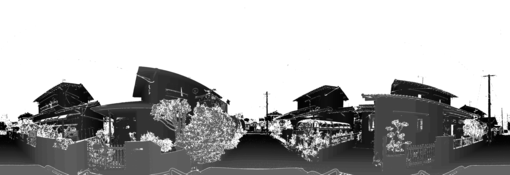,width= 0.45\columnwidth}
        & \psfig{file=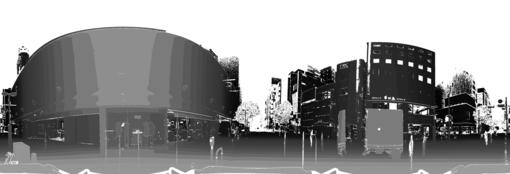,width= 0.45\columnwidth}  \\
        \psfig{file=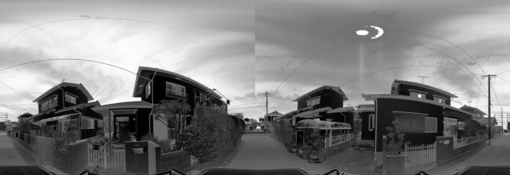,width= 0.45\columnwidth}
        & \psfig{file=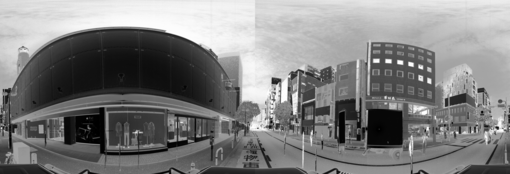,width= 0.45\columnwidth}  \\
        \psfig{file=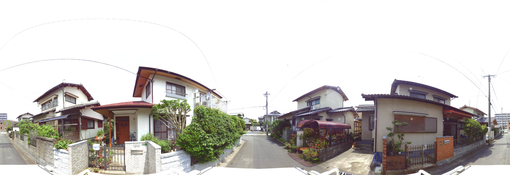,width= 0.45\columnwidth}         
        & \psfig{file=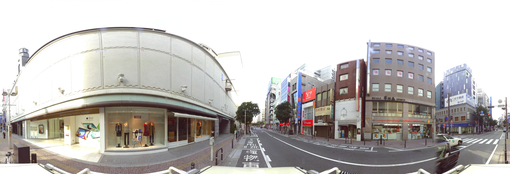,width= 0.45\columnwidth}    \\ \hline

    \end{tabular}

\caption{\label{fig:dataset_faro} \small
  \href{http://robotics.ait.kyushu-u.ac.jp/~kurazume/research-e.php?content=db-hidden}{\it Dense MPO Dataset}: examples of high-resolution range, reflectance and color panoramic images for six outdoor place categories: forest, coast, indoor/outdoor parking lot, residential and urban area. In range images, darker colors indicate closer distances and in reflectance images, brighter colors indicate higher intensity.}
\end{figure}

The dataset is aimed to be used for visual outdoor place categorization in a mobile robot and autonomous vehicle platform. We took scans in six different place categories: `forest', `coast', `indoor parking lot', `outdoor parking lot', `residential area' and `urban area'. For example, while recording ``urban area" place categories, we located a laser scanner on a straight road, a corner and an intersection road in different urban area. We labelled a ground truth of place category for each scan of the proposed datasets. Some examples of synchronized panoramic images for six place categories are shown in Fig.~\ref{fig:dataset_faro}. 

The proposed dataset contains 103$\sim$112 scans for each category and 650 number of scans in total. As shown in Table~\ref{tab:faro}, each category is consisted of seven different area and each particular area contains 12$\sim$17 locations of scans. For each scan, we moved the car 10-100 meters, stopped and took the full panoramic scan and color images. We repeated this process several times in each area. A point of a laser scan and a pixel of color images were synchronized afterwards using the SCENE software. 

%http://robotics.ait.kyushu-u.ac.jp/~kurazume/research-e.php?content=db-hidden

\subsubsection{Data format} For each scan we provide 4 different multi-modal information: color 3D point cloud, color panoramic image, reflectance panoramic image and range panoramic image. 

\begin{itemize}

\item{\bf Point cloud data} is consist of 3D point cloud, which is X, Y, and Z coordinates in general. In this dataset, we scanned outdoor place scan of color 3D point cloud with 360$^{\circ}$ fields of view from a laser scanner. Since the laser scanner can measure the time of flight for a laser pulse and the power of the laser reflectivity from the object, the 3D point cloud consists of a synchronized color 3D point with reflectance, e.g., X, Y, Z, reflectance, R, G and B in order.

\item{\bf Panoramic image} is an stitched image that is horizontally elongated fields of view. In this research, we create a scan that covers 360$^{\circ}$ horizontal fields of view. In order to create a 2D scan image, a 3D point data can be transformed into a range and projected as a range panoramic image. In the same manner, the reflectivity can be represented as reflectance and color panoramic image. 

\end{itemize}

%%%%%%%%%%%%%%%%%%%%%%%%%%%%%%%%%%%%%%%%%%%%%%%%%%%%%%%%%%%%%%%%%%%%%%%%%%%%%%%%

\section{SPARSE MPO DATASETS}

The \href{http://robotics.ait.kyushu-u.ac.jp/~kurazume/research-e.php?content=db-hidden}{\it Sparse MPO Dataset}~\cite{MPO2016} is composed of a low-resolution of (32 $\times$ 2166 pixels) panoramic scan and a color 360$^{\circ}$ spherical image and a GPS location using several sensors. The aim of proposing the second dataset is to utilize in real-time dynamic platform, such as a mobile robot and an autonomous vehicle. The total number of consecutive frame is 34,200 scans with six place categories which are same as the \href{http://robotics.ait.kyushu-u.ac.jp/~kurazume/research-e.php?content=db-hidden}{\it Dense MPO Dataset}. In this section, we will explain about sensors, experimental setup, data description and a map of \href{http://robotics.ait.kyushu-u.ac.jp/~kurazume/research-e.php?content=db-hidden}{\it Sparse MPO Dataset}. 

% Although the measuring time of each laser scanner differs from each other, the resolution of the scan data is completely different. In {\it Sparse MPO Dataset}, Velodyne laser scanner can capture a low-resolution scan, but with a 10Hz high frequency of update rate. While moving the mobile robot, Velodyne can capture its surrounding environment in real-time. The Velodyne dataset contains approximately 5,700 pairs of scans that belong to one specific category and the total scans of the dataset are 34,200 pairs, as presented in Table~\ref{tab:velo}. 

%\begin{figure}[thpb]
%\centering
%%
%    \begin{tabular}{cc}
%        %
%        \psfig{file=Figures/experiment_car.png,width= 0.34\columnwidth} &
%        \psfig{file=Figures/experiment.jpg,width= 0.57\columnwidth} 
%        %
%    \end{tabular}
%%
%\caption{\label{fig:experiment} \small
%  Experimental setup for {\it Sparse MPO Dataset} equipped with a Velodyne HDL-32E laser scanner, a Kodak PIXPRO SP360 camera and a GARMIN GPS 18x LVC}
%\end{figure}

\subsection{Data acquisition}    %DATA ACQUISITION
In this dataset, we used several sensor modalities, such as a Velodyne HDL-32E laser scanner, a Kodak PIXPRO SP360 camera and a GARMIN GPS 18x LVC. The sensor includes a time-of-flight laser scanner and a camera, so that it can provide range, reflectance and color panoramic images. The sensor system is installed on top of vehicle with 2 meters height. 

\subsubsection{Velodyne HDL-32E} 
A laser scanner can perform with a maximum range of 70 meters. The LiDAR provided the field of view of $41.3^{\circ} (-30.67^{\circ} \sim 10.67^{\circ}) \times 360^{\circ}$ (vertical $\times$ horizontal). Each pair of a complete panoramic scan shows a resolution of 32 $\times$ 2166 points and the horizontal and vertical angular resolution are 0.17$^{\circ}$ and 1.33$^{\circ}$, respectively. Velodyne rotates the laser beam around the vertical axis with maximum update rate in 10 Hz, which obtain complete panoramic range and reflectance data, but in this experiment, we reduced the update rate in 2 Hz. 

\subsubsection{Kodak PIXPRO SP360} 
A camera can record 360$^{\circ}$ view angle of 16.36 Megapixels image with 360$^{\circ}$ spherical curved lens. In this experiment, we record a color spheric image in 6$\sim$7 Hz next to the Velodyne laser scanner.

\subsubsection{GARMIN GPS 18x LVC} 
A GPS is WAAS-enabled GPS receiver with 12-parallel-channel. It provides a pulse-per-second logic-level output with a rising edge aligned to within 1 microsecond of UTC time. The output data is recorded NMEA 0183 format in 2 Hz.

\subsubsection{Experimental setup}
In the \href{http://robotics.ait.kyushu-u.ac.jp/~kurazume/research-e.php?content=db-hidden}{\it Sparse MPO Dataset}, we installed a group of sensors on top of a mobile platform with 2 meters height. A laser scanner rotates around the vertical axis to obtain a 3D point cloud around the mobile robot. Next to the laser scanner, we placed a camera and GPS, as we shown in Fig.~\ref{fig:experiment}.

Since the measuring time of laser scanner shows a large difference from the previous dataset, the scanning process and a resolution of scan are completely different. The previous FARO Focus3D system can capture a high-resolution of color 3D point clouds, however it took 3 minute per a single scan. On other other hand, Velodyne HDL-32E laser scanner can take 0.2 sec per a single scan since it obtains a relatively low-resolution of scan.

\begin{table}[t]
\centering
\footnotesize
\caption{
\small
SPARSE MPO DATASET of outdoor scene containing 34,200 pairs of range and reflectance scans, GPS position data and color images
}
\label{tab:velo}
\begin{tabular}{|c|c|c|c|c|c|c|c|c|c|c|c|}    
    \hline
    \multirow{3}{*}{Category}    & \multicolumn{5}{c|}{Number of scans by location} & \multirow{3}{*}{Total} \\ \cline{2-6}
       & Set1    & Set2    & Set3    & Set4    & Set5 &  \\ \cline{2-6}
       & Set6    & Set7    & Set8    & Set9    & Set10 & \\ \hline    
    \multirow{2}{*}{Coast}      & 511    & 254    & 571    & 221    & 314 & \multirow{2}{*}{4298} \\ \cline{2-6}
       & 376    & 872    & 506    & 386    & 287    &    \\ \hline
    \multirow{2}{*}{Forest}     & 440    & 824     & 980     & 707    & 730 & \multirow{2}{*}{6479}     \\ \cline{2-6}
       & 720    & 439    & 311    & 797    & 531    &    \\ \hline
    \multirow{2}{*}{Parking lot (in)}      & 520     & 357     & 274     & 873     & 583    & \multirow{2}{*}{4780}     \\ \cline{2-6}
        & 343     & 466     & 592     & 344     & 428     &    \\ \hline
    \multirow{2}{*}{Parking lot (out)}      & 874     & 579     & 388     & 370     & 477    & \multirow{2}{*}{5445}     \\ \cline{2-6}
        & 536     & 581     & 563     & 460     & 617     &    \\ \hline 
    \multirow{2}{*}{Residential area} & 674     & 787     & 667     & 724     & 563    & \multirow{2}{*}{7464}        \\ \cline{2-6}
      & 973     & 717     & 720     & 977     & 662     &    \\ \hline    
    \multirow{2}{*}{Urban area}  & 490    & 572    & 587    & 487    & 410  & \multirow{2}{*}{5734}    \\ \cline{2-6}
        & 566    & 712    & 565    & 606    & 739    &    \\ \hline    
    \multirow{2}{*}{Total scans}    & 3509    & 3373    & 3467    & 3382    & 3077  & \multirow{2}{*}{34200}       \\ \cline{2-6}
     & 3514    & 3787    & 3257    & 3570    & 3264    &  \\ \hline
\end{tabular}
% \end{center}
\vspace{-3mm}
\end{table}

\subsection{Dataset descriptions}    % DATA DESCRIPTIONS

\subsubsection{Place category}
The \href{http://robotics.ait.kyushu-u.ac.jp/~kurazume/research-e.php?content=db-hidden}{\it Sparse MPO Dataset} is aimed for visual place categorization in outdoor environments, like the \href{http://robotics.ait.kyushu-u.ac.jp/~kurazume/research-e.php?content=db-hidden}{\it Dense MPO Dataset}. As we shown in Table~\ref{tab:velo}, we took the six different place categories: `forest', `coast', `indoor parking lot', `outdoor parking lot', `residential area' and `urban area'. We labelled a ground truth of place category for each scan of the proposed datasets. Examples of scans and the complete panoramic images are shown in Fig.~\ref{fig:dataset_faro}. 

\begin{figure}[thpb]
\centering
    \begin{tabular}{|c|}     \hline
        {\cellcolor[HTML]{EFEFEF}Range and reflectance panoramic images}    \\ \hline
        
        Forest \\
        \psfig{file=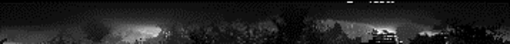,width= 0.9\columnwidth}  \\
        \psfig{file=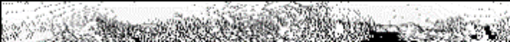,width= 0.9\columnwidth}  \\ \hline
        Coast \\ 
        \psfig{file=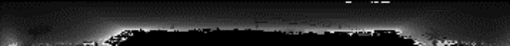,width= 0.9\columnwidth}  \\
        \psfig{file=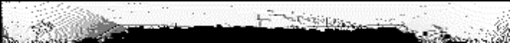,width= 0.9\columnwidth}  \\ \hline
        Indoor parking lot \\
        \psfig{file=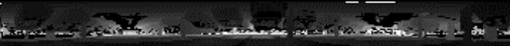,width= 0.9\columnwidth}  \\
        \psfig{file=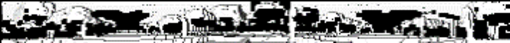,width= 0.9\columnwidth}  \\ \hline
        Outdoor parking lot \\
        \psfig{file=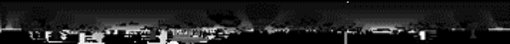,width= 0.9\columnwidth}  \\
        \psfig{file=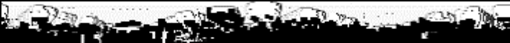,width= 0.9\columnwidth}  \\ \hline
        Residential area   \\ 
        \psfig{file=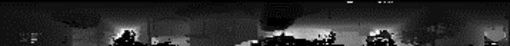,width= 0.9\columnwidth}  \\
        \psfig{file=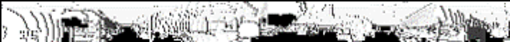,width= 0.9\columnwidth}  \\ \hline
        Urban area  \\
        \psfig{file=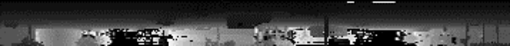,width= 0.9\columnwidth}  \\
        \psfig{file=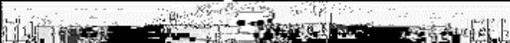,width= 0.9\columnwidth}  \\ \hline
    \end{tabular}
\caption{\label{fig:dataset_velo} \small
  \href{http://robotics.ait.kyushu-u.ac.jp/~kurazume/research-e.php?content=db-hidden}{\it Sparse MPO Dataset}: examples of low-resolution range and reflectance panoramic images for six outdoor place categories: `forest', `coast', `indoor parking lot', `outdoor parking lot', `residential area' and `urban area'. In range images, darker colors indicate closer distances and in reflectance images, brighter colors indicate higher intensity.}
\end{figure}

\subsubsection{Location map}
The \href{http://robotics.ait.kyushu-u.ac.jp/~kurazume/research-e.php?content=db-hidden}{\it Sparse MPO Dataset} contains approximately 5,700 pairs of scans that belong to one specific category and the total scans of the dataset are 34,200 pairs in Fukuoka City, Japan, as we presented in Table~\ref{tab:velo}. All the locations of the dataset's scan are presented as a map of \href{http://robotics.ait.kyushu-u.ac.jp/~kurazume/research-e.php?content=db-hidden}{\it Sparse MPO Dataset}, as shown in Fig.~\ref{fig:map_velo}. 
We took our dataset while driving the mobile robot with 30$\sim$50 kph velocity and 2$\sim$7 Hz sensor data in real-time.

\section{PLACE CATEGORIZATION}

In our previous research for indoor place categorization~\cite{mozos2012sensors,mozos2013ar,jung2016local} we applied local features to extract the patterns of visual place camera image or 3D scan data for indoor environment. Local binary patterns(LBP) and its extensions are well-performed with 2D images. Multi-modal local feature discriminates 3D indoor place datasets that are captured by RGB-D sensor or laser scanner. In this section, place categorization procedures are introduced.
%However, the 3D place categorization for outdoor environment is difficult to evaluate, since the 3D outdoor place dataset exists very few. One of the main reason is that 3D outdoor data is expensive for time consumption and equipment cost to capture a single scan. RGB-D sensor can be an well-known approach for indoor 3D place dataset, but the limitation is a short detectable range and a weakness of strong sunlight. In this research, we categorize 3D outdoor places created by a laser scanner. 

\subsection{Feature extraction}

\subsubsection{Local binary patterns}
The main purpose of local binary patterns (LBP) operator~\cite{ojala2002pami} is to describe the relationships between the values of center pixels and their neighboring pixels. For example, using a single image modality of local binary transformation of each pixel $\mathit{i = (x, y)}$ in an image $\mathit{I}$, let $\mathcal{N}_{P}{(i)}$ be each pixel's $P$-neighborhood. I compare each value $\mathit{I}(i)$ of the image at position $(x_i,y_i)$ with each of the values ${s}_{LBP}(z)$ corresponding to the neighboring pixels $\{{j}_{0}, {j}_{1},\cdots,{j}_{P-1}\}\in \mathcal{N}_{P}{(i)}$. For each comparison, a binary value ${s}_{LBP}(z) \in \{0, 1\}$ indicates that the value $\mathit{I}(i)$ is bigger or smaller than the neighboring pixel's value $\mathcal{N}_{P}{(i)}$:

\boldmath
\begin{equation}
\mathit{I}_{LBP}(i) = \sum_{k=0}^{P-1}{s}_{LBP}(\mathit{I}(i)-\mathit{I}(j_k)){2}^{k},
\forall {j_k} \in \mathcal{N}_{P}(i),	 
\end{equation}
\unboldmath

\begin{equation}
{s}_{LBP}(z)= \left\{  \begin{array}{ll}
         1 & \mbox{if ${z} \geq 0$ }\\
         0 & \mbox{if ${z} < 0$ } \end{array}
\right. 	\\
	%\qquad 
\label{equ:LBPdef}
\end{equation}

\noindent where ${s}_{LBP}(z)$ indicates the binary decision of a single digit ${2}^{k}$ by comparing the reference pixel $\mathit{i}$ and the neighboring pixel $\mathit{j}$ of a modality $z$. The resulting binary values are ordered clockwise starting with the value which is to the right of the reference pixel ${j}_{0}$ at position $(x_i+1,y_i)$. The corresponding binary string is then transformed into decimal value $d$ in the range $[0,\cdots,255]$. This decimal value $d$ will be the new value for pixel $i$.

%\begin{figure}[thpb]
%%
%\centerline{
%\hfill
%\psfig{file=Figures/lbp.png,width=\columnwidth} 
%\hfill
%}
%%
%\caption{\label{fig:lbp} \small An example of local binary transformation of the center pixel using 8-neighborhood ($P=8$) for a range image ${I}_{Dep}$: 
%   The left image presents an example of pixels from a range image. The middle image presents a center pixel (marked in red) and the corresponding eight neighboring pixels (marked in black). The right image shows the binary results (0 marked in black and 1 marked in white) of the comparisons and the resulting decimal value (marked in red).
%}
%%
%\end{figure}

%The resulting binary values, LBP = ${(01100010)}_2$ = 98, are ordered clockwise starting with the value which is to the right of the reference pixel ${j}_{0}$ at position $(x_i+1,y_i)$. The corresponding binary string is then transformed into decimal value $d$ in the range $[0,\cdots,255]$. This decimal value $d$ will be the new value for pixel $i$. ]An example of this process is shown in Fig.~\ref{fig:lbp}.
%In this research, a single modality that the range panoramic image $\mathit{I}_{Dep}$ is utilized. The measurement is applied to the local transformation and can be transformed in each resulting LBP image $I_{LBP_{Dep}}$.

\subsubsection{Spin image}
The spin image~\cite{johnson1998surface} is one of popular technique for surface matching and 3D object recognition. The conventional spin images encode the global properties of any surface in an object-oriented coordinate system. However, in our place categorization, we applied spin image in a scanner-oriented view point in cylindrical coordinate system rather than in an object-oriented view point using 3D surface of object. A entire 3D point cloud can be represented as the scanner-oriented image with the cylindrical coordinate system when the position of laser scanner is (0,0,0) for each scan data. To achieve this we use the tangent plane through oriented perpendicularly to n and the line through parallel to~\cite{johnson1998surface}.

%Conventionally, the important element of spin image generation is to select oriented points of 3D surface points. Although we use the scan oriented coordinate system, our oriented points is the position of laser scanner which is (0,0,0) for each scan data. The surface normal is upright direction of laser scanner. With defined, we can formulate a 2D basis which will correspond to a local coordinate system. To achieve this we use the tangent plane through oriented perpendicularly to n and the line through parallel to~\cite{johnson1998surface}. This will result in a cylindrical coordinate system where is the (non-negative) perpendicular distance to while
%is the signed (positive or negative) perpendicular distance to~\cite{johnson1998surface}.
%
%- more equation and details of spin image -

\subsubsection{Texton}
A popular texture analysis named Texton~\cite{Leung:2001:RRV:543015.543017} is a filter-based technique. The Leung-Malik (LM) filter bank and the maximum response (MR) filter bank contains filters at multiple orientations and scales. In standard texton, the maximum response filter among several Gaussian and LOG (Laplacian of Gaussian) filters with the same scale, but different orientations are selected. In this experiment, we adopted the maximum response filter among filters with different scales but same orientation so as to evaluate the directions of edges, not the size of them. 

%The MR8 filter bank consists of 38 filters but only 8 filter responses. The filter bank contains filters at multiple orientations but their outputs are ''collapsed'' by recording only the maximum filter response across all orientations. This achieves rotation invariance. The filter bank is shown in figure 4 and consists of a Gaussian and a Laplacian of Gaussian (these filters have rotational symmetry), an edge filter at 3 scales and a bar filter at the same 3 scales. The latter two filters are oriented and occur at 6 orientations at each scale. Measuring only the maximum response across orientations reduces the number of responses from 38 (6 orientations at 3 scales for 2 oriented filters, plus 2 isotropic) to 8 (3 scales for 2 filters, plus 2 isotropic). Detailed information about the filters can be found in section 3.2.

\subsection{Histogram of feature}
The transformed feature image $I$ is presented by a histogram $\bf{h}$ of length $l$ in which each bin ${\bf{h}}\unboldmath(l)$ indicates the frequency of appearance of the descriptor's pattern $d$. For example, the LBP image $I_{\mathrm{LBP}}$ is then represented by a histogram $\bf{{h}_{LBP}}$ as

\begin{equation}
{\bf{h}_{LBP}}\unboldmath(l) = \sum_{i} \mathcal{I}(\mathit{I}_{LBP}(i)=d) , 
\label{eq:hist_lbp}
\end{equation}

\noindent where $\mathcal{I}$ denotes the indicator function which returns $1$ if its argument is true, and $0$ otherwise. Since the LBP values are restricted to the $8$-neighborhood, the dimension of the final histogram is $l=256$. Each bin indicates the frequency of appearance of each decimal value $d \in [0,\cdots,255]$. 

%In the same manner, the spin image $I_{\mathrm{Spin}}$ is then represented by a histogram $\bf{{h}_{Spin}}$ as follow
%
%\begin{equation}
%{\bf{h}_{Spin}}\unboldmath(l) = \sum_{i} \mathcal{I}(\mathit{I}_{Spin}(i)=d) .
%\label{eq:hist_spin}
%\end{equation}

%In the LBP feature descriptor, we define any out-of-range and non-returnable values obtained by using a laser scanner as a Not-A-Number ($NAN$) value. Therefore, the additional $NAN$ value to a histogram bin is not assigned as point values. The final histogram of feauture $\bf{{H}_{LBP}}$ indicates a concatenation of the histogram $\bf{{h}_{LBP}}$ and the a Not-A-Number ($NAN$) value as
% 
%\begin{equation}
%{\bf{H}_{LBP}}\unboldmath = \big\{{\bf{h}_{LBP}}\unboldmath, NAN\big\}.
%\end{equation}
%
%\noindent Since the additional $NAN$ value is assigned distinctively, the total numbers of LBP histogram bins are $2^{8}+1=257$. 

\subsection{Classification}
We use Support vector machines (SVM)~\cite{Cortes1995ml,bishop2006book} with a radial basis function (RBF) kernel for the categorization. Multi-class classification is performed by the ``one-against-one" approach~\cite{Knerr2010springer}. In our experiments, we use the LIBSVM library~\cite{chang2011libsvm}. Following the method reported in~\cite{Wei2010svm}, the parameters $C$ and $\gamma$ are selected by a grid search using cross-validation. The ranges of $C$ and $\gamma$ are $C \in [2^{-1},\cdots, 2^{20} ]$ and $\gamma \in [2^{-20} ,\cdots, 2^{0}]$ in the grid search.

\subsection{Majority vote technique}
We apply a classic linear-time majority vote technique~\cite{boyer1991mjrty} for consecutive frames of \href{http://robotics.ait.kyushu-u.ac.jp/~kurazume/research-e.php?content=db-hidden}{\it Sparse MPO Dataset}. When a classification result using SVM ${C_{t}}$ at the time $t$, we can define $M$ number of consecutive frames $\{{C_{t}},{C_{t-1}},\cdots,{C_{t-M}}\}$ as an element. As following equation~\ref{eq:mvote}, We find a majority of frames ${C_{t}}^{Final}$ in the element

\begin{equation}
    {C_{t}}^{Final}=\text{Majority vote}({C_{t}},{C_{t-1}},\cdots,{C_{t-M}}).
\label{eq:mvote}
\end{equation}

\noindent In this experiment, a boundary between different places and an ambiguous area visually is effective for the majority vote technique. %, in Fig.~\cite{fig:mvote}.

%\begin{figure}[thpb]
%%
%\centerline{
%\hfill
%\psfig{file=Figures/majorityvote.png,width=\columnwidth} 
%\hfill
%}
%%
%\caption{\label{fig:mvote} \small An example of majority vote technique in {\it Sparse MPO Dataset}
%}
%%
%\end{figure}

\section{EXPERIMENTAL EVALUATION}

As our preliminary work, we conducted several place categorization experiments using two outdoor place datasets: \href{http://robotics.ait.kyushu-u.ac.jp/~kurazume/research-e.php?content=db-hidden}{\it Dense MPO Dataset} and \href{http://robotics.ait.kyushu-u.ac.jp/~kurazume/research-e.php?content=db-hidden}{\it Sparse MPO Dataset}. Firstly, we evaluate the performance of several standard feature descriptors (e.g., LBP, Spin image, Texton and LNP) with supervised learning method, SVM. We compare the performance of the majority vote technique for \href{http://robotics.ait.kyushu-u.ac.jp/~kurazume/research-e.php?content=db-hidden}{\it Sparse MPO Dataset} additioinally.

\subsection{Dense MPO Datasets}

We present classification results of \href{http://robotics.ait.kyushu-u.ac.jp/~kurazume/research-e.php?content=db-hidden}{\it Dense MPO Dataset} by comparing the conventional feature descriptor, LBP, Spin image and Texton. As we shown in Table~\ref{tab:comp_dense}, in a single range image, LBP shows better performance with 94.35\% of correct classification ratio (CCR), while Spin image and Texton descriptors show 89.43\% and 89.04\% respectively. Additionally, LBP feature using reflectance panoramic image shows a bit higher 96.43\% CCR and the confusion matrix is shown in Table~\ref{tab:conf_lbp_dense}.

To extend the our multi-modalities, we evaluate the reflectance and color panoramic image by applying LBP and Texton. Among the multi-modalities, LBP descriptor with reflectance modality shows the best performance 96.42\% in overall. In addition, Table~\ref{tab:comp_dense_m} shows results of LBP and LTP~\cite{jung2016local} for utilizing the multiple modalities, and those are performed over 90\%.

\begin{table}
    \centering
    \caption{Correct classification ratio (CCR) results of standard descriptors with single modality using DENSE MPO DATASET [\%]}        
		\begin{tabular}{|c|c|c|c|}
		  \hline
		  Descriptor & Range & Reflectance & Color \\
		  \hline
		  Spin image~\cite{johnson1998surface} & $89.43\pm0.00$  &  -   &  - \\   
		  \hline
		  LBP~\cite{ojala2002pami}   & $94.35\pm2.67$  & $\mathbf{96.42\pm2.68}$  & $93.86\pm3.85$ \\  
		  \hline
		  Texton~\cite{Leung:2001:RRV:543015.543017} & $89.04\pm5.58$  & $73.59\pm13.73$  & $81.81\pm10.67$\\   
		  \hline
%		  \multirow{2}{*}{Multiple} & LBP & \multicolumn{2}{c|}{$95.67\pm3.69$}  &  -  \\   
%		  \cline{2-5}
%		  & LTP & \multicolumn{2}{c|}{$92.84\pm3.33$}   &  - \\   
%		  \hline

		\end{tabular}
\label{tab:comp_dense}
\end{table}

\begin{table}[t]
\tiny
\caption{Confusion matrix of reflectance LBP images for DENSE MPO DATASET [CCR\%]}
\centering
\begin{tabular}{|c|c|c|c|c|c|c|}
\cline{2-7}
\multicolumn{1}{c|}{} & \multicolumn{1}{c|}{Coast} & \multicolumn{1}{c|}{Forest} & \multicolumn{1}{c|}{ParkingIn} & \multicolumn{1}{c|}{ParkingOut} & \multicolumn{1}{c|}{Residential} & \multicolumn{1}{c|}{Urban} \\ \hline
Coast  & \cellcolor[HTML]{C0C0C0}$\mathbf{93.45\pm17.39}$ & ${6.54\pm17.39}$ & ${0.00\pm0.00}$ & ${0.00\pm0.00}$ & ${0.00\pm0.00}$ & ${0.00\pm0.00}$  \\ \hline % \cline{2-8} 
Forest    & ${8.95\pm8.27}$ & \cellcolor[HTML]{C0C0C0}$\mathbf{91.04\pm8.27}$ & ${0.00\pm0.00}$ & ${0.00\pm0.00}$ & ${0.00\pm0.00}$ & ${0.00\pm0.00}$  \\ \hline % \cline{2-8}
ParkingIn & ${0.00\pm0.00}$ & ${0.00\pm0.00}$ & \cellcolor[HTML]{C0C0C0}$\mathbf{100.00\pm0.00}$ & ${0.00\pm0.00}$ & ${0.00\pm0.00}$  & ${0.00\pm0.00}$   \\ \hline % \cline{2-8}
ParkingOut & ${1.91\pm2.92}$ & ${0.00\pm0.00}$ & ${0.00\pm0.00}$ & \cellcolor[HTML]{C0C0C0}$\mathbf{98.08\pm2.92}$ & ${0.00\pm0.00}$  & ${0.00\pm0.00}$   \\ \hline % \cline{2-8}
Residential & ${1.33\pm2.68}$ & ${0.00\pm0.00}$ & ${0.00\pm0.00}$ & ${0.00\pm0.00}$ & \cellcolor[HTML]{C0C0C0}$\mathbf{97.32\pm4.43}$ & ${1.33\pm4.00}$   \\ \hline % \cline{2-8}
Urban & ${0.00\pm0.00}$ & ${0.00\pm0.00}$ & ${0.00\pm0.00}$ & ${0.00\pm0.00}$ & ${1.87\pm4.00}$ & \cellcolor[HTML]{C0C0C0}$\mathbf{98.12\pm4.00}$   \\ \hline % \cline{2-8} 
\end{tabular}
\label{tab:conf_lbp_dense}
\end{table}

\begin{table}
    \centering
    \caption{Correct classification ratio (CCR) results of multi-modalities using DENSE MPO DATASET}        
		\begin{tabular}{|c|c|c|}
		  \hline
		  Modality & \multicolumn{2}{c|}{Range + Reflectance} \\ \hline
		  Descriptor & LBP~\cite{mozos2012sensors} & LTP~\cite{jung2016local} \\ \hline
		  CCR[\%] &    $95.67\pm3.69$ & $92.84\pm3.33$ \\
		  \hline
		\end{tabular}
\label{tab:comp_dense_m}
\end{table}

\subsection{Sparse MPO Dataset}

We evaluate categorization of {\it Sparse MPO Dataset} by comparing the conventional feature descriptor, LBP and Spin image. Using a range panoramic image, LBP also shows better performance with 83.98\% of correct classification ratio (CCR) and the spin image shows 79.23\% in Table~\ref{tab:comp_velo_vote}. Additionally, we applied both feature descriptor for majority vote technique that means finding the majority of time-series data. Then, spin image and LBP are increased the 6$\sim$10\% CCR results.  

\begin{figure}[thpb]
\centering
    \psfig{file=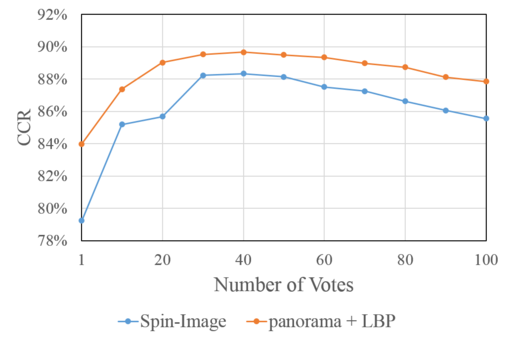,width=\columnwidth}
\caption{\label{fig:consec_vote} \small
  A performance of {\it Sparse MPO Dataset} by applying a majority vote technique}
\end{figure}

For the majority vote technique, we compare a numbers of vote for finding the majority of predictions using the $M$ previous frames. The highest performance was $M=40$ numbers of vote with 20 seconds long streaming scan data and 88.34\% and 89.67\% for each spin image and LBP descriptor respectively, as shown in Fig.~\ref{fig:consec_vote}. A confusion matrix of LBP with majority vote is presented in Table~\ref{tab:conf_lbp_vote}.

\begin{table}[t]
    \centering
    \caption{Correct classification ratio (CCR) results of standard descriptors and majority vote using SPARSE MPO DATASET [\%]}        
		\begin{tabular}{|c|c|c|}
		  \hline
		  \multirow{2}{*}{Descriptor} & \multicolumn{2}{c|}{Technique} \\
		  \cline{2-3}	  
		      & None & Majority vote \\
		  \hline
		  Spin image~\cite{johnson1998surface} & $79.23\pm4.51$  & $88.34\pm0.12$\\   
		  \hline
		  LBP~\cite{ojala2002pami} & $83.98\pm4.59$  & $\mathbf{89.67\pm0.21}$\\   
		  \hline
		\end{tabular}
\label{tab:comp_velo_vote}
\end{table}

\begin{table}[t]
\tiny
\caption{Confusion matrix of LBP images with majority vote for SPARSE MPO DATASET [CCR\%]}
\centering
\begin{tabular}{|c|c|c|c|c|c|c|}
\cline{2-7}
\multicolumn{1}{c|}{} & \multicolumn{1}{c|}{Coast} & \multicolumn{1}{c|}{Forest} & \multicolumn{1}{c|}{ParkingIn} & \multicolumn{1}{c|}{ParkingOut} & \multicolumn{1}{c|}{Residential} & \multicolumn{1}{c|}{Urban} \\ \hline
Coast  & \cellcolor[HTML]{C0C0C0}$\mathbf{82.99\pm4.19}$ & ${5.20\pm0.54}$ & ${0.00\pm0.00}$ & ${7.15\pm2.20}$ & ${1.50\pm0.05}$ & ${3.16\pm0.56}$  \\ \hline % \cline{2-8} 
Forest    & ${8.64\pm1.12}$ & \cellcolor[HTML]{C0C0C0}$\mathbf{90.87\pm1.20}$ & ${0.00\pm0.00}$ & ${0.49\pm0.01}$ & ${0.00\pm0.00}$ & ${0.00\pm0.00}$  \\ \hline % \cline{2-8}
ParkingIn & ${0.00\pm0.00}$ & ${3.31\pm0.01}$ & \cellcolor[HTML]{C0C0C0}$\mathbf{93.95\pm0.65}$ & ${2.16\pm0.56}$ & ${0.59\pm0.02}$  & ${0.00\pm0.00}$   \\ \hline % \cline{2-8}
ParkingOut & ${0.96\pm0.06}$ & ${0.00\pm0.00}$ & ${5.25\pm0.12}$ & \cellcolor[HTML]{C0C0C0}$\mathbf{89.16\pm1.31}$ & ${2.44\pm0.26}$  & ${2.19\pm0.26}$   \\ \hline % \cline{2-8}
Residential & ${0.45\pm0.01}$ & ${0.00\pm0.00}$ & ${0.00\pm0.00}$ & ${2.70\pm0.01}$ & \cellcolor[HTML]{C0C0C0}$\mathbf{93.23\pm0.20}$ & ${3.63\pm0.20}$   \\ \hline % \cline{2-8}
Urban & ${0.00\pm0.00}$ & ${0.00\pm0.00}$ & ${0.00\pm0.00}$ & ${3.84\pm0.76}$ & ${8.76\pm0.79}$ & \cellcolor[HTML]{C0C0C0}$\mathbf{87.39\pm0.20}$   \\ \hline % \cline{2-8} 
\end{tabular}
\label{tab:conf_lbp_vote}
\end{table}

\section{CONCLUSIONS}
We have proposed \href{http://robotics.ait.kyushu-u.ac.jp/~kurazume/research-e.php?content=db-hidden}{\it Dense and Sparse MPO Datasets} for benchmarking outdoor place categorization. The datasets consist of six outdoor places categories and 650 color 3D point cloud with reflectance for \href{http://robotics.ait.kyushu-u.ac.jp/~kurazume/research-e.php?content=db-hidden}{\it Dense MPO Dataset} and 34,200 3D point clouds with reflectance, a spherical image and a GPS location for \href{http://robotics.ait.kyushu-u.ac.jp/~kurazume/research-e.php?content=db-hidden}{\it Sparse MPO Dataset}. For benchmarking, we presented some of standard classification results.

%
%We have presented {\it Dense 3D Kyushu University Outdoor Place Datasets} and {\it Sparse 3D Kyushu University Outdoor Place Datasets} for outdoor place categorization. %The category of outdoor semantic places consisted of forest, residential area, parking lot and urban area with around 35 locations per each category. The outdoor SICK and FARO LiDAR dataset included two pairs of 143 images and three pairs of 140 images, in total. The high resolution LiDAR dataset is very useful to benchmark vision-based semantic place labeling in outdoor environments.  In the future we plan on expanding the set of available dataset by adding additional 3D scans and color images labels for currently unlabeled new categories. Furthermore, we plan on evaluating our benchmark dataset using various feature descriptors, such as GIST, SIFT, LBP and LNP, with various resolution of scans images.

\addtolength{\textheight}{-12cm}   % This command serves to balance the column lengths
                                  % on the last page of the document manually. It shortens
                                  % the textheight of the last page by a suitable amount.
                                  % This command does not take effect until the next page
                                  % so it should come on the page before the last. Make
                                  % sure that you do not shorten the textheight too much.

%%%%%%%%%%%%%%%%%%%%%%%%%%%%%%%%%%%%%%%%%%%%%%%%%%%%%%%%%%%%%%%%%%%%%%%%%%%%%%%%

%%%%%%%%%%%%%%%%%%%%%%%%%%%%%%%%%%%%%%%%%%%%%%%%%%%%%%%%%%%%%%%%%%%%%%%%%%%%%%%%

%%%%%%%%%%%%%%%%%%%%%%%%%%%%%%%%%%%%%%%%%%%%%%%%%%%%%%%%%%%%%%%%%%%%%%%%%%%%%%%%
%\section*{APPENDIX}

%Appendixes should appear before the acknowledgment.

%%%%%%%%%%%%%%%%%%%%%%%%%%%%%%%%%%%%%%%%%%%%%%%%%%%%%%%%%%%%%%%%%%%%%%%%%%%%%%%%
%\section*{ACKNOWLEDGMENT}

%The preferred spelling of the word ÒacknowledgmentÓ in America is without an ÒeÓ after the ÒgÓ. Avoid the stilted expression, ÒOne of us (R. B. G.) thanks . . .Ó  Instead, try ÒR. B. G. thanksÓ. Put sponsor acknowledgments in the unnumbered footnote on the first page.

%%%%%%%%%%%%%%%%%%%%%%%%%%%%%%%%%%%%%%%%%%%%%%%%%%%%%%%%%%%%%%%%%%%%%%%%%%%%%%%%

\bibliographystyle{ieeetr}
\bibliography{jung2016iros}

\end{document}